\documentclass[letterpaper, 10 pt, conference]{ieeeconf}
\IEEEoverridecommandlockouts
                                                          
\usepackage{amsmath,amssymb,amsfonts}
\usepackage[pdftex]{graphicx}
\usepackage{textcomp}
\usepackage{xcolor}
\usepackage{hyperref}
\usepackage{amsmath}
\usepackage[noend]{algpseudocode}
\usepackage{comment}
\usepackage[normalem]{ulem}
\usepackage[colorinlistoftodos]{todonotes}
\usepackage[framemethod=tikz]{mdframed}
\newtheorem{dfn}{Definition}

\def\BibTeX{{\rm B\kern-.05em{\sc i\kern-.025em b}\kern-.08em
    T\kern-.1667em\lower.7ex\hbox{E}\kern-.125emX}}

\usepackage{graphicx}
\usepackage{psfrag}
\usepackage{pstool}
\usepackage{multirow}
\usepackage{soul}
\usepackage{gensymb}
\algnewcommand\algorithmicforeach{\textbf{for each}}
\algdef{S}[FOR]{ForEach}[1]{\algorithmicforeach\ #1\ \algorithmicdo}

\newcommand{\usefootref}[1]{%
    $^{\ref{#1}}$%
}

\DeclareUnicodeCharacter{0301}{\'{e}}

\usepackage{xspace}

\makeatletter

\DeclareRobustCommand\onedot{\futurelet\@let@token\@onedot}
\def\@onedot{\ifx\@let@token.\else.\null\fi\xspace}

\usepackage{algorithm,algpseudocode,setspace}

\DeclareMathOperator*{\argmax}{arg\,max}

\usepackage{etoolbox}
\makeatletter
\patchcmd{\@makecaption}
  {\scshape}
  {}
  {}
  {}
\makeatletter
\patchcmd{\@makecaption}
  {\\}
  {.\ }
  {}
  {}
\makeatother
\def\tablename{Table}

\font\bfmath=cmmib10
\mathchardef\Gamma="7100
\mathchardef\Delta="7101
\mathchardef\Theta="7102
\mathchardef\Lambda="7103
\mathchardef\Xi="7104
\mathchardef\Pi="7105
\mathchardef\Sigma="7106
\mathchardef\Upsilon="7107
\mathchardef\Phi="7108
\mathchardef\Psi="7109
\mathchardef\Omega="710A

\mathchardef\alpha="710B
\mathchardef\beta="710C
\mathchardef\gamma="710D
\mathchardef\delta="710E
\mathchardef\epsilon="710F
\mathchardef\zeta="7110
\mathchardef\eta="7111
\mathchardef\theta="7112
\mathchardef\iota="7113
\mathchardef\kappa="7114
\mathchardef\lambda="7115
\mathchardef\mu="7116
\mathchardef\nu="7117
\mathchardef\xi="7118
\mathchardef\pi="7119
\mathchardef\rho="711A
\mathchardef\sigma="711B
\mathchardef\tau="711C
\mathchardef\upsilon="711D
\mathchardef\phi="711E
\mathchardef\chi="711F
\mathchardef\psi="7120
\mathchardef\omega="7121
\mathchardef\epsilon="7122

\mathchardef\varepsilon="7122
\mathchardef\vartheta="7123
\mathchardef\varpi="7124
\mathchardef\varrho="7125
\mathchardef\varsigma="7126
\mathchardef\varphi="7127
\mathchardef\imath="717B
\mathchardef\jmath="717C

\def\smallbfW{{\raise1.5pt\hbox{\mbox{\boldmath $_W$}}}}

\let\ts=\thinspace

\def\my4psfrag#1#2#3#4#5#6#7#8{
        \begin{figure}[htp]
        \begin{center}
            \begin{tabular}[h]{c c}
              {\leavevmode{\includegraphics[width=#1truecm]{#2.eps}}}
              &
              {\leavevmode{\includegraphics[width=#1truecm]{#3.eps}}} \\
              {\leavevmode{\includegraphics[width=#1truecm]{#4.eps}}}
              &
              {\leavevmode{\includegraphics[width=#1truecm]{#5.eps}}}
         \end{tabular}
           \vspace{#6}
           \caption{#7}
           \label{#8}
        \end{center}
        \end{figure}
}

\def\mydouble4psfrag#1#2#3#4#5#6#7#8{
        \begin{figure*}[htp]
        \begin{center}
            \begin{tabular}[h]{c c}
              {\leavevmode{\includegraphics[width=#1truecm]{#2.eps}}}
              &
              {\leavevmode{\includegraphics[width=#1truecm]{#3.eps}}} \\
              {\leavevmode{\includegraphics[width=#1truecm]{#4.eps}}}
              &
              {\leavevmode{\includegraphics[width=#1truecm]{#5.eps}}}
         \end{tabular}
           \vspace{#6}
           \caption{#7}
           \label{#8}
        \end{center}
        \end{figure*}
}

\usepackage[style=ieee,doi=false,mincitenames=1,maxcitenames=20,minbibnames=1,maxbibnames=20,isbn=false,url=true]{biblatex}
\makeatother
\usepackage[belowskip=-7pt,aboveskip=10pt,font={small}]{caption}
\usepackage[font={small}]{subcaption}
\makeatletter

\begin{document}

\title{ 
Ensemble Latent Space Roadmap for Improved Robustness \\in Visual Action Planning

}

\author{Martina Lippi*$^{1}$, Michael C. Welle*$^{2}$, Andrea Gasparri$^{1}$, Danica Kragic$^{2}$
\thanks{*These authors contributed equally (listed in alphabetical order).}
\thanks{ ${}^1$Roma Tre University, Rome, Italy  {\it\small \{martina.lippi,andrea.gasparri\}@uniroma3.it} }%
\thanks{ ${}^2$KTH Royal Institute of Technology Stockholm, Sweden, {\it\small \{mwelle,dani\}@kth.se}}%
}

\maketitle

\begin{abstract}

 Planning in learned latent spaces helps to decrease the dimensionality of raw observations.
In this work, we propose to leverage the ensemble paradigm to 
enhance the robustness of latent planning systems. We rely on our Latent Space Roadmap (LSR) framework, which builds a graph in a learned structured latent space to perform planning.  
Given multiple LSR  framework instances, that differ either on their latent spaces or on the parameters for constructing the graph, we use the action information as well as the embedded nodes of the produced plans to define similarity measures. These are then utilized to select the most promising plans. We validate the performance of our Ensemble LSR (ENS-LSR) on simulated box stacking and grape harvesting tasks as well as on a real-world robotic T-shirt folding experiment. 
\end{abstract}

\section{Introduction}
\label{sec:intro}

Generating plans from raw observations can be more flexible than relying on handcrafted state extractors or modeling the system explicitly.
This can be particularly advantageous in complex tasks, such as manipulating deformable objects or operating in dynamic environments, where accurately defining the underlying system state can prove to be challenging.
Moreover, the generation of plans from raw observations offers the potential to create both visual and action plans, which we refer to as visual action plans. The visual plan captures the sequence of images that the system transitions through while executing the action plan.  This  allows human operators to easily comprehend the  plan and gain a better understanding of the system behavior.

Nonetheless, 
the high-dimensionality of raw observations 
challenges the effectiveness of classical planning approaches~\cite{bellman1961curse}. In this regard, representation learning  may help to retrieve compact representations from high-dimensional observations. Given these representations, classical planning approaches can be then employed. Based on this principle, we proposed in our previous works \cite{lippi2022enabling, lippi2022augment} a Latent Space Roadmap (LSR) framework.  
This involves training a deep neural network to learn a low-dimensional latent space of the high-dimensional observation space. 
A roadmap is then constructed within this space to capture the connectivity between different states, and is used to perform planning.

\begin{figure}[t!]
    \centering
    \includegraphics[width=\linewidth]{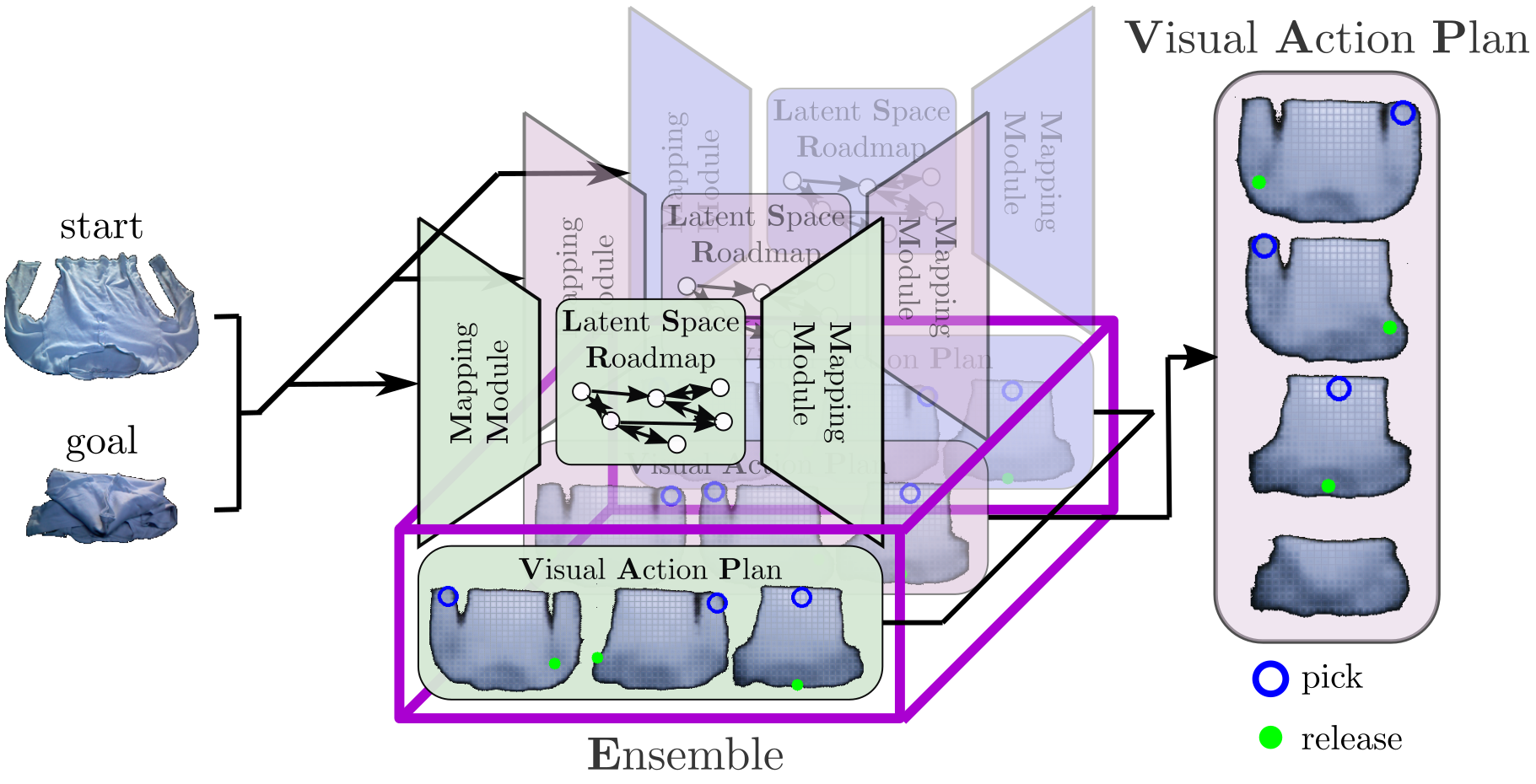}
    \vspace{-15pt}
    \caption{Depiction of the Ensemble Latent Space Roadmap framework for a folding task. Given start and goal observations, multiple models produce different visual action plans, and the ones with the highest degree of similarity to one another are selected. 
}
    \label{fig:en_overview}
\end{figure}

 While utilizing learned models for representation is generally advantageous, they might lead to unreliable plans. For instance, this can occur when working with non-representative data samples during the training process.  
In this work, we aim to enhance the robustness of latent space planning by  incorporating 
the  ensemble paradigm \cite{rokach2019ensemble}: 
several models are collected and their plans are combined based on similarity measures. 
 To this aim, we 
 take into account both the sequence of actions and the composition of transited nodes in the latent space.
 In line with majority voting ensemble approaches \cite{ruta2005classifier}, we select the plans which are the most similar to the others.
We name the resulting framework Ensemble LSR (ENS-LSR) and show an overview in Fig.~\ref{fig:en_overview}.
We demonstrate its effectiveness  compared to \cite{lippi2022enabling, lippi2022augment} in simulated box stacking and grape harvesting tasks and on a real-world folding task. 
In detail, our contributions are:
\begin{itemize}
    \item The design of a novel ensemble algorithm for latent space planning along with the definition of appropriate similarity measures. 
    \item An extensive performance comparison on multiple tasks, including a new simulated agricultural task and a real-world manipulation task with deformable objects.
    \item Extensive ablation studies analyzing different components of the algorithm. 
    \item The release of all datasets, code, and execution videos for real-world experiments on the project website\footnote{\label{fn:website} \url{https://visual-action-planning.github.io/ens-lsr/}}.
\end{itemize}
Note that, although we focus on LSR framework, the proposed algorithm can be  adapted to any latent space planning method.

\section{Related Work} \label{sec:rw}

\subsection{Visual planning}
Visual planning  allows robots to generate plans and make decisions based on the visual information they perceive from their environment. 
Methods directly working in the image space have been proposed in the literature. For instance, the method in \cite{finn2017deep} is based on Long-Short Term Memory blocks to generate a video prediction model, which is  then integrated into a Model Predictive Control (MPC) framework to produce  visual plans; the approach in \cite{wang2019learning} produces visual foresight plans for rope manipulation using GAN models and then resorts to a learned rope inverse dynamics. 
However, as mentioned in the Introduction, latent spaces can be employed to reduce the high-dimensionality of the image space. For instance, an RRT-based algorithm in the latent space with collision checking  is adopted in \cite{Ichter2019}; interaction features conditioned on object images are learned and integrated within Logic-Geometric Programming for planning in \cite{Toussaint_ral2022};  model-free Reinforcement Learning (RL) in the off-policy case combined with auto-encoders is investigated in~\cite{yarats2021improving}. 
Despite the dimensionality reduction of the space, 
the approaches above typically require a large amount of data to operate. In contrast, the LSR framework is able to accomplish visual planning in a data-efficient manner by leveraging a contrastive loss to structure the latent space.   

\subsection{Ensemble methods}
While the ensemble principle has primarily been utilized within the machine learning community \cite{dong2020survey}, it has also been implemented in robotic planning to combine multiple planning algorithms or models and improve the overall performance of the system.
Multiple planners in parallel,  which work under diverse assumptions, are executed according to the method in  \cite{Scherer_ICRA2015}. An ensemble selection is then made based on learned priors on planning performance. The combination of multiple heuristics in greedy best-first search planning algorithm is investigated in \cite{Helmert_2021}. 
An online ensemble learning method based on different predictors is proposed instead in \cite{Zambelli_TCDS2017} to achieve an accurate robot forward model that can be beneficial for imitation behavior.   
The study in \cite{Adil_Access2020} adopts 
an ensemble framework with neural networks to mitigate the error in stereo vision systems and then performs planning for manipulation. 
 Finally, several applications of the ensemble paradigm to RL approaches can be found in the literature, such as in 
\cite{Wiering_TSMC2008,buckman2018sample,lee2021sunrise}. In detail, the ensemble method in \cite{Wiering_TSMC2008} combines the value functions of five different RL algorithms, the approach in \cite{buckman2018sample}  dynamically interpolates between rollouts with different horizon lengths, while the study in \cite{lee2021sunrise} proposes ensemble-based weighted Bellman backups. 
However, to the best of our knowledge, none of the above approaches is able to realize visual action planning.

\section{Preliminaries and Problem Formulation} \label{sec:proplemdef}

\subsection{Visual Action Planning} 
Let $\mathcal{O}$ be the set of all  possible observations, and $\mathcal{U}$ the set of all feasible actions of the system.
\begin{dfn}\label{defn::vap}
Given start and goal observations $O_s$ and $O_g$, respectively,  a Visual Action Plan (VAP)  $P=(P^o,P^u)$ consists of a visual plan $P^o$ and action plan $P^u$, where the visual plan $P^o=(O_s=O_0,O_1,...,O_n=O_g)$, with $O\in \mathcal{O}$, contains a sequence of $n$ observations showing intermediate states from start to goal, while the action plan $P^u=(u_0,u_1,...,u_{n-1})$ provides  the  respective actions to transition through the states, with $u\in \mathcal{U}$.
\end{dfn}

Note that in general, given start and goal observations, 
there are many potential plans that can feasibly transition the system from start to goal, i.e., VAPs may not be unique.

\subsection{Latent Space Roadmap-based system}\label{sec:SLSR}
The Latent Space Roadmap framework, presented 
in \cite{lippi2022enabling}, is an approach to realize visual action planning. 
The core idea of the method is to map the observations in a low dimensional structured latent space  $\mathcal{Z}$ where a roadmap is built to perform planning. 
More specifically, the latent space is structured to reflect the separation of the underlying states of the system, i.e., latent states associated to the same underlying system state are clustered together, while latent states associated to different system states are far apart; 
based on this latent space, a roadmap captures the possible transitions among the clusters and allows to find plans in the latent space. The plans generated in the latent space are then decoded to obtain visual plans. 
\begin{dfn}\label{defn::lsr}
    A Latent Space Roadmap-based system  \mbox{S-LSR$=\{\xi,\mathcal{G},\omega\} $} consists of a latent mapping function $\xi$, that maps observations to a low  dimensional structured latent space, $\xi:\mathcal{O} \to \mathcal{Z}$, a Latent Space Roadmap   $\mathcal{G}=\{\mathcal{V},\mathcal{E} \}$, that is a graph structure composed of the sets of nodes
    $\mathcal{V}$,  and edges  
    $\mathcal{E}$, and an observation generator function $\omega$, that maps latent representations $z\in \mathcal{Z}$ into observations, $\omega:\mathcal{Z}\to \mathcal{O}$.
\end{dfn}
We enclose the two functions $\xi$ and $\omega$ in a single module, which we refer to as Mapping Module (MM). 
 Note that by either changing the mapping module or the LSR we can obtain different S-LSR models, where the generic S-LSR$_i$ is defined as S-LSR$_i=\{\xi_i,\mathcal{G}_i,\omega_i\}$.
  For example, three S-LSRs are depicted in different colors in Fig. \ref{fig:en_overview}.

\begin{figure*}
    \centering
    \includegraphics[width=\textwidth]{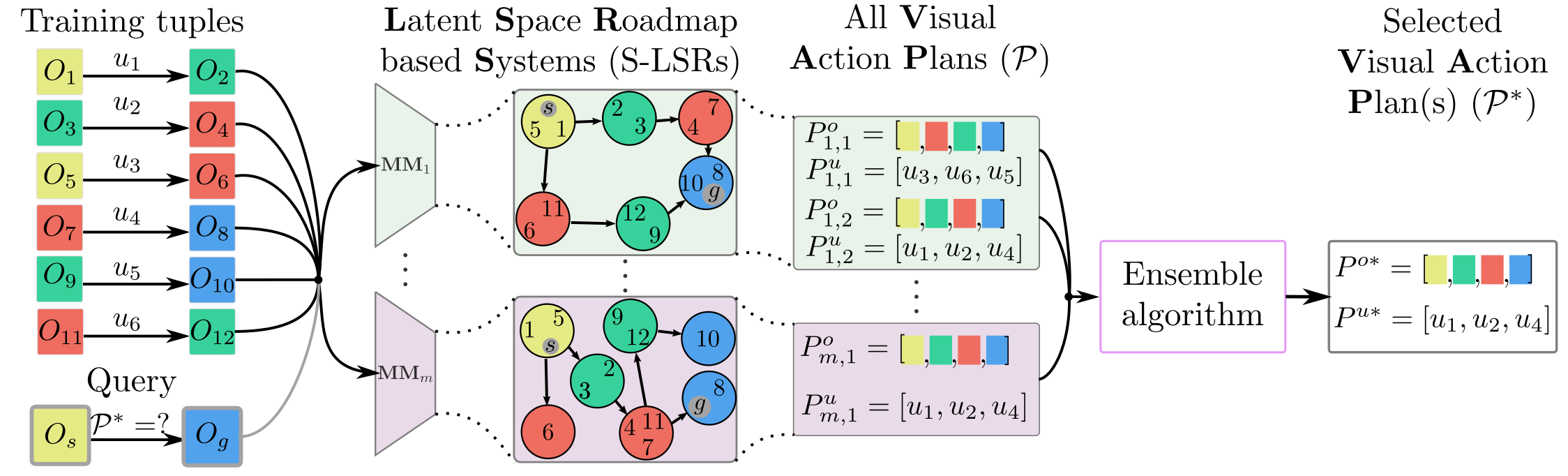}
    \caption{Overview of our ensemble approach. The training tuples (with $a=1$) are mapped through different mapping modules into separate latent spaces, where individual LSRs are  built. Colors of the observations represent their underlying state.
    Given start and goal observations (bottom left), these are mapped into the latent spaces (grey circles) and all the possible visual action plans are extracted from the $m$ S-LSRs. Finally, a selection is made by the ensemble algorithm. }

    \label{fig:ensemble_explained}
\end{figure*}

\noindent
\textbf{Dataset and S-LSR building:} The S-LSR assumes a dataset $\mathcal{T}_o$ consisting of tuples $(O_i,O_j,\rho)$ where $O_i$ and $O_j$ are two observations and $\rho$ represents the action information between them.
In detail, 
$\rho=\{a,u\}$ comprises a binary indicator variable $a\in \{0,1\}$, which indicates if an action has taken place ($a=1$) or not ($a=0$), along with the respective action specification $u$, which is meaningful only in case $a=1$. 
This representation allows us to also model  variations in observations not caused by actions, such as changes in lighting conditions. For such task irrelevant factors of variation it holds $a=0$, i.e., no action took place, and the value of the variable $u$ is ignored. 
 Without loss of generality, we consider that the observations in the dataset are numbered and we refer to
  $i$ in $O_i$ as its index. 
Note that the dataset does not require any knowledge of the underlying system states in the observations, but only the information of the action  between them.

Briefly, based on the above dataset structure, the mapping functions $\xi$ and $\omega$ are realized with an  encoder-decoder based architecture. A contrastive loss term is introduced which exploits the action binary indicator variable: 
states in tuples with $a=1$ are encouraged to be spaced in the latent space, while states with $a=0$ are encouraged to be close\cite{lippi2022enabling}. 
The LSR $\mathcal{G}$  is then built in the latent space by clustering and connecting the latent states associated with the training dataset: each cluster is associated with a node, and edges are constructed if actions exist in the dataset to transition among the respective nodes. 
 For each node, we consider as its representative latent state the cluster centroid. 
Furthermore, according to the Action Averaging Baseline method from \cite{lippi2022enabling}, each edge is  endowed with the average action among all the actions between the respective nodes of the edge. 
According to the procedure in \cite[Algorithm 2]{lippi2022enabling}, the LSR building mainly  depends on a single parameter, $c_{\max}$,  that is  an upper bound on the number of weakly connected components of the graph. Intuitively, we build the LSR by maximizing the number of edges, in order to increase likelihood of finding plans,  while constraining the number of possible  disconnected components of the graph, in order to avoid fragmentation of the graph.

\noindent
\textbf{Planning procedure:} 
Given a S-LSR and the start $O_s$ and goal $O_g$ observations, visual action planning is achieved by first mapping the given  start and end observations  into the structured latent space $\mathcal{Z}$ via the latent mapping function $\xi$. Then, the closest nodes in the graph $\mathcal{G}$ with  centroids in $z_s$ and $z_g$, respectively, are retrieved and the, possibly multiple, shortest paths in  $\mathcal{G}$  from $z_s$ to $z_g$ are computed. These paths provide both the sequences of latent states from start to goal, referred to as latent plans \mbox{$P^z=(z_s=z_0,z_1,...,z_n=z_g)$},  and the sequences of respective actions  $P^u$,  retrieved  from the graph edges $\mathcal{E}$. The latent plans are finally decoded into visual plans $P^o$ through the observation generator $\omega$. 
We re-iterate that multiple shortest paths can be found from $z_s$ to $z_g$ and denote by $q$ the number of these paths.

\subsection{Problem Formulation}

Given a dataset $\mathcal{T}_o$, the correctness of plans suggested by a S-LSR mainly depends on two factors \cite{ccpaper}: \emph{i)} how well the latent space is structured, i.e., if states are properly separated, and \emph{ii)} how well the LSR is built, i.e., if nodes are 
representative of the underlying states of the system as well as if they are correctly connected according to the possible actions of the system. 
Both points are influenced by the choice of the respective hyperparameters, i.e., given the same dataset and same start and goal observations, two S-LSRs may provide distinct plans, which also differ in terms of correctness, depending on the used hyperparameters.   Among the hyperparameters, we also include the constitution of the overall dataset into validation and training parts. 
Our objective is to design a robust system that can perform visual action planning while mitigating high variance in the results arising with different parameters of the S-LSR.

\section{Ensemble LSR Algorithm}\label{sec:enslsr}

To solve the above problem, we propose to exploit an \emph{ensemble} of $m$ S-LSRs that, as a whole, allow for robustness against outliers. We refer to the ensemble model as ENS-LSR, i.e.,  $\text{ENS-LSR}=\{\text{S-LSR}_1,...,\text{S-LSR}_m\}$.  
The basic idea is that,  given a start and a goal observation, different visual action plans are returned by the S-LSRs and  a selection is made among them based on their similarity. 

Let $\mathcal{P}_i$ be the set of VAPs produced by the S-LSR~$i$ given start $O_s$ and goal $O_g$ observations, i.e.,  
$$\mathcal{P}_i = \{P_{i,1}, ...\,, P_{i,q_i}\},$$ 
where $P_{i,j}$ is the VAP $j$ generated by the S-LSR and $q_i$ is the  number of total VAPs. We define the set of all potential VAPs, from $O_s$ to $O_g$, associated with the ensemble model as the union of the sets $\mathcal{P}_i$, $\forall i\in\{1,...,m\}$, i.e., 
$$\mathcal{P}=\mathcal{P}_1\cup \mathcal{P}_2 \cup ... \cup \mathcal{P}_m.$$ 
Our objective is to select the subset of VAPs $\mathcal{P}^*\subseteq\mathcal{P}$ to provide as output by the ensemble model.

An illustrative overview of the method is provided in Fig.~\ref{fig:ensemble_explained} where multiple S-LSRs composing the ensemble model are depicted. The colors of the observations represent their underlying states, e.g., observations with indices $1$ and $5$ represent the same state  in this setting.
The figure shows how, starting from  the same training tuples, different latent spaces are produced by the mapping modules in ENS-LSR and different LSRs are built. Therefore,  
given a start and a goal observation (bottom left), different visual action plans are returned by the S-LSRs. At this point, a selection is made by the ensemble algorithm among these plans based on their similarity. 
More specifically, in line with majority voting rule approaches in ensemble learning \cite{ruta2005classifier}, we propose to select the VAPs which are the most similar to each other, thus removing outlier ones. This implies that, as first step,   appropriate similarity measures must be defined. Then, an algorithm is designed to select the paths based on these measures.
 In the reminder of the section, we address these two steps.  
Note that a naive ensemble approach could be to simply propose all plans suggested by all models in ENS-LSR, i.e., $\mathcal{P}\equiv \mathcal{P}^*$.  However, in this way no outlier rejection is made and wrong paths can be possibly provided as output 
(for more details see the results in Sec.~\ref{sec:exp}).

\subsection{Similarity measures}\label{sec:sim-measure}
We identify two main ways to evaluate the similarity between VAPs: \emph{i)} with respect to actions in the action plans, 
and \emph{ii)} with respect to node compositions in the latent plans.

\noindent
\textbf{Action similarity}
As recalled in Sec. \ref{sec:SLSR}, the action plans are obtained by  concatenating  the average actions contained in the edges of the respective latent plans. 
Let $P_{i,j}^u$ be the action plan of the $j$~th  VAP produced by the S-LSR$_i$. 
We evaluate the similarity between two action plans $P_{i,j}^u$ and $P_{k,l}^u$ by computing the cosine similarity  which captures if the vectors point towards the same direction. 
To compute this similarity, we preprocess the action plans as follows: first, in order to compare all the coordinates of the actions, we  collapse the actions of the plans in one dimension, then, in order to compare plans with different lengths, we pad the shortest with zeros to equal the lengths of the sequences.  For instance, in case of a 2D pick and place action, the action variable can be written as 
$u = (p, r)$, where $p = (p_x, p_y)$ and $r = (r_x, r_y)$ represent the pick and release positions, and its collapsed version is given by $\bar{u} = (p_x, p_y, r_x,r_y)$. We denote the vector obtained by preprocessing the action plan $P_{i,j}^u$ as $\bar{P}^u_{i,j}$. Therefore, the action similarity measure $s^u$ is obtained as 
\begin{equation}\label{eq:ac-measure}
    s^u( \bar P_{i,j}^u,\bar P_{k,l}^u):=\frac{1}{2}\left(1+\frac{\bar P_{i,j}^u\cdot \bar P_{k,l}^u}{\|\bar P_{i,j}^u  \|\,\|\bar P_{k,l}^u\|}\right),
\end{equation}
 with $i\neq k$, which is in the range $[0,1]$ and reaches $1$ when the plans are the same, and $0$ when they are maximally dissimilar, i.e., they point in opposite directions.  
Further baselines for the action similarity measure, such as the euclidean and the  edit distances,  have been validated in the Sec.~\ref{sec:exp}, which are outperformed by \eqref{eq:ac-measure}.

\noindent
\textbf{Node similarity}
As recalled in Sec. \ref{sec:SLSR}, the nodes of the $i$\ts th LSR are obtained by clustering the latent states associated with the training observations. We endow each node with the set of indices of the respective training observations and refer to it as node \emph{composition}. For instance, in Figure~\ref{fig:ensemble_explained}, the node compositions of the two green nodes in the  top LSR are $\{2,3\}$ and $\{12,9\}$.   Clearly, 
if two LSRs  have similar node compositions, they will likely produce plans in agreement. 
Let $P_{i,j}^z$ be the latent plan $j$ of S-LSR$_i$ and $\bar P_{i,j}^z$ be the collection of indices associated with the latent plan, i.e., it is the union of the compositions of the nodes in the plan.  
We define the node similarity measure $s^n$ for two latent plans $P_{i,j}^z$ and ${P}^z_{k,l}$ as their Jaccard similarity, i.e., as the ratio of the cardinality of the intersection of the two sets to the cardinality of the union:  
\begin{equation}\label{eq:node-measure}
    s^n( \bar P_{i,j}^z,\bar P_{k,l}^z):= \frac{|\bar P_{i,j}^z\cap \bar P_{k,l}^z|}{|\bar P_{i,j}^z\cup \bar P_{k,l}^z|},
\end{equation}
with $i\neq k$ and $|\cdot|$ the cardinality of the set $(\cdot)$. This measure is in the range $[0,1]$ and reaches $1$ when the plans are the same 
and $0$ when they are maximally dissimilar  (no intersection among the plans).
Note that the above measure can deal with plans having different lengths by construction. 

\vspace{0.2cm}
We define the overall similarity measure $s$ between two VAPs $P_{i,j}$ and $P_{k,l}$ as the sum of the 
action and the node similarity measures, i.e., $s = s^u+ s^n$. Note that both the measures $s^u$ and $s^n$ are in the interval $[0,1]$,  making them comparable. 

\subsection{Ensemble Latent Space Roadmap Algorithm}
Algorithm \ref{alg::ensemble} shows the proposed Ensemble LSR algorithm. It takes as input the set of $m$ S-LSRs composing the ensemble ENS-LSR, as well as the start $O_s$ and goal $O_g$ observations, 
and returns the most suited visual action plan(s), i.e., $\mathcal{P}^*$, from all the possible ones, i.e., $\mathcal{P}$. The basic idea is to compute a \emph{cumulative comparison score} $c_{i,j}$ for each VAP $P_{i,j}$, based on both action and node similarity measures with respect to the other VAPs, and subsequently select the ones with highest scores. In particular, the score $c_{i,j}$ is obtained by considering, for each S-LSR$_k$ with $k\neq i$, the respective VAP with highest overall similarity measure to $P_{i,j}$ and then by summing up all these highest measures $\forall k\in\{1,...,m\}, k\neq i$. Note that, for each S-LSR$_k$, with $k\neq i$, we consider only one VAP in the computation of the score to ensure that all S-LSRs have equal relevance in that calculation, i.e., to prevent that S-LSRs with many VAPs than others have greater influence in the score than the latter.

 At start, the set $\mathcal{P}$ composed of all VAPs, from $O_s$ to $O_g$, produced by the $m$ S-LSRs is computed (line \ref{lst:line:vaps}) and   an empty set $\mathcal{C}$ is initialized (line \ref{lst:line:ec}) to store the cumulative comparison scores. 
Then, the algorithm iterates over the VAPs of each $\text{S-LSR}_i$ (line \ref{lst:line:lsri}-\ref{lst:line:pij}). 
For each $P_{i,j}$,  the corresponding action $P_{i,j}^u$ and latent $P_{i,j}^z$ plans are preprocessed as previously described (lines \ref{lst:line:ai_p} and \ref{lst:line:ni_p}) and the cumulative comparison score $c_{i,j}$ is initialized to zero  (line \ref{lst:line:cinit}). 
 At this point, a comparison is made  with the plans produced by the other S-LSRs. 
Therefore, an iteration is made over the sets of VAPs $\mathcal{P}_k$, with $k\neq i$, obtained by the other S-LSRs and, for each $k$, an empty set $\mathcal{S}$ is initialized to store the similarity measures between $P_{i,j}$ and the VAPs in $\mathcal{P}_k$. For each VAP $P_{k,l}$  in $\mathcal{P}_k$, 
the corresponding preprocessed action $\bar P_{k,l}^u$ and latent $\bar P_{k,l}^z$ plans are obtained (lines \ref{lst:line:ak_p}-\ref{lst:line:nk_p}) and their action $s^u$ and node $s^n$ similarities with the plans  $\bar P_{i,j}^u$ and $\bar P_{i,j}^z$ are computed (lines \ref{lst:line:ac-sim}-\ref{lst:line:node-sim}) according to \eqref{eq:ac-measure} and \eqref{eq:node-measure}, respectively. The overall similarity between the VAPs $P_{i,j}$ and $P_{k,l}$ is calculated by summing the the individual similarities $s^u$ and $s^n$  and is stored in the set $\mathcal{S}$  (line \ref{lst:line:addtos}).  Once all the overall similarities with the VAPs of the S-LSR$_k$ have been computed, the maximum one is added  to the comparison score variable $c_{i,j}$ (line \ref{lst:line:cumscore}). The latter score is complete when all S-LSRs different from $i$ have been analyzed, and is then added to the set $\mathcal{C}$  (line \ref{lst:line:addtoc}).

Once  all the cumulative comparison scores have been computed, the set of indices $\mathcal{I}$ of the VAPs with highest scores is extracted (line \ref{lst:line:is}), i.e., 
$$\mathcal{I}^*  = \argmax_{i,j} \,\mathcal{C}, $$
and the corresponding VAPs $\mathcal{P}^*$ are selected as final output  (line \ref{lst:line:ps}).

\begin{algorithm} \caption{Ensemble LSR Planning}
\small
\def\negsp{\vspace{-5pt}}
\def\negup{\vspace{-7pt}}
\def\negin{\vspace{-3pt}}
\setstretch{1.2}
\begin{algorithmic}[section]
\Require  $\text{ENS-LSR} = \{\text{S-LSR}_1, ..., \text{S-LSR}_m\}$, $O_s$, $O_g$
    \begin{algorithmic}[1]
    \State $\mathcal{P}\leftarrow $ Compute-VAPs (ENS-LSR, $O_s, O_g$) \label{lst:line:vaps}
    \State $\mathcal{C} \leftarrow \{ \}$ \label{lst:line:ec}
    \ForEach{$\mathcal{P}_i \in \mathcal{P} $} \label{lst:line:lsri}
        \ForEach{$P_{i,j} \in \mathcal{P}_i$}\label{lst:line:pij}
            \State $\bar P_{i,j}^u \leftarrow $ Preprocess-action-plan($P_{i,j}$) \label{lst:line:ai_p}
            \State $\bar P_{i,j}^z \leftarrow $ Preprocess-latent-plan($P_{i,j}$) \label{lst:line:ni_p}
            \State $c_{i,j} \leftarrow 0$ \label{lst:line:cinit}
            \ForEach{$\mathcal{P}_k \in \mathcal{P}\text{, with } k\neq i$} \label{lst:line:lsrj}
                    \State $\mathcal{S} = \{ \}$ \label{lst:line:ecp}          
                    \ForEach{$P_{k,l} \in \mathcal{P}_k$}
                        \State $\bar P_{k,l}^u \leftarrow $ Preprocess-action-plan($P_{k,l}$) \label{lst:line:ak_p}
                        \State $\bar P_{k,l}^z \leftarrow $ Preprocess-latent-plan($P_{k,l}$) \label{lst:line:nk_p}
                        \State $s^u \leftarrow $ Action-sim($\bar P_{i,j}^u, \bar P_{k,l}^u $)$\quad$ [eq. \eqref{eq:ac-measure}]\label{lst:line:ac-sim}
                        \State $s^n \leftarrow $ Node-sim($\bar P_{i,j}^z, \bar P_{k,l}^z $) $\quad\,$ [eq. \eqref{eq:node-measure}] \label{lst:line:node-sim}
                        
                        \State $\mathcal{S} \leftarrow  \mathcal{S} \cup \{s^u + s^n$\} \label{lst:line:addtos}
                    \EndFor
                    \State $c_{i,j}\leftarrow c_{i,j} + \max(\mathcal{S})$ \label{lst:line:cumscore}
            \EndFor   
            \State $\mathcal{C}\leftarrow\mathcal{C}\cup \{c_{i,j}\}$ \label{lst:line:addtoc}
        \EndFor
    \EndFor
    \State $\mathcal{I}^* \leftarrow\argmax_{i,j}(\mathcal{C})$\label{lst:line:is}
    \State $\mathcal{P}^* \leftarrow \text{Select-VAPs}(\mathcal{P},\mathcal{I}^*)$\label{lst:line:ps}
   
    \end{algorithmic}
\Return $\mathcal{P}^*$
\end{algorithmic}
\label{alg::ensemble}
\end{algorithm}

\section{Simulation results}\label{sec:exp}
\begin{figure}
    \centering
    \includegraphics[width=0.7\linewidth]{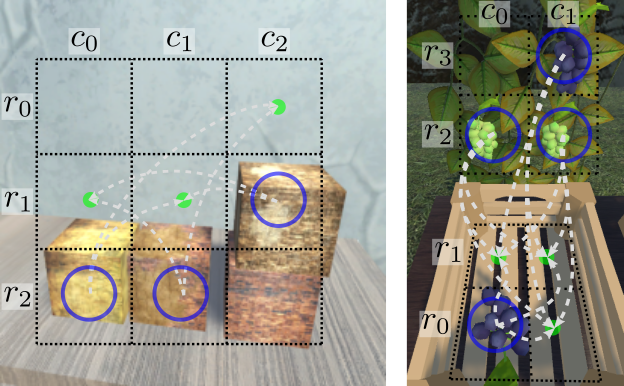}
\vspace{-1pt}
    \caption{Illustrations of the simulation setups: box stacking task on the left and grape harvesting on the right. Possible actions are highlighted with arrows from the picking positions (blue circles) to the release ones (green dots). 
    }
    \label{fig:simu_setup}
\end{figure}

We validate our ensemble LSR algorithm on two simulation tasks realized with Unity3D engine \cite{unitygameengine}: \emph{i)} a box stacking task, previously introduced in \cite{lippi2022enabling} (where it is referred to as hard box stacking task) and shown in Fig. \ref{fig:simu_setup} on the left, as well as a grape harvesting task, shown in Fig. \ref{fig:simu_setup} on the right. Actions in both tasks are expressed as pick and place operations, i.e., $u=(p,r)$ as reported in Sec.~\ref{sec:sim-measure}. 
 The datasets consist of $2500$ and $5000$ tuples for stacking and harvesting, respectively, and are available on the project website\usefootref{fn:website}.
The box stacking task consists of four different boxes, with similar textures, that can be stacked in a $3\times3$ grid. Varying lighting conditions and noise (up to $15\%$) in the box positioning are introduced in the observations as task irrelevant factors of variation. 
The following stacking rules apply: \emph{i)} only one box can be moved at the time, \emph{ii)} boxes cannot float or be placed outside of the grid, \emph{iii)}
only one box can be placed in a single cell, and \emph{iv)} a box can be moved only if the top cells are empty. All the actions are reversible in the task, i.e., if an action $u=(p,r)$ exists from observation $O_i$ to $O_j$, it follows that the transition from $O_j$ to $O_i$ using the reversed action \mbox{$u'=(r,p)$} is possible. 
This implies that the bidirectional edges in the LSRs are built.

The grape harvesting task on the other hand consists of eight cells. Four cells are located in the box and the other four cells are on the vine. The scenario represents a small-scale harvesting task with four grapes, two of the white variety and two of the black one. 
As task irrelevant factors of variation, we introduce different lighting conditions, different scales of the bunches ($\pm 10\%$), positional noise inside the cells ($\pm 12.5\%$), different orientation inside the box ($\pm 180\degree $), and the bunch model is randomly chosen (with probability equal to $0.5$) between two that differ in number of grapes. 
The rules for the task are the following: \emph{i)} only one bunch can be moved at the time, \emph{ii)} only one bunch per cell and \emph{iii)} bunches can only be placed inside the box, i.e., a bunch cannot be moved from the box to the vine or from vine to vine. The last rule makes visual action planning for this task significantly more challenging compared to the box stacking task as it implies that actions are not reversible and the LSRs are directed graphs. Furthermore, several irrelevant factors of variations are introduced in this task compared to the box stacking one. Finally, it also highlights the broad  field of application of ENS-LSR, which is here deployed in an agricultural setting as in the  European  project CANOPIES.

In both tasks the system state is given by the arrangement of the objects in the cells. 
These two simulation setups are used to answer the following questions: 
\begin{enumerate}
    \item What is the improvement provided by ENS-LSR compared to individual S-LSRs with different mapping modules and how does the number $m$ of S-LSRs  affect the performance?
    \item Can ENS-LSR simplify the tuning of the LSR building parameter $c_{\max}$?
    \item What is the ensemble performance when alternative similarity measures are employed and what is the contribution of the individual similarity measures? 
\end{enumerate}

To evaluate the performance, 
 we use the same evaluation setting as in our previous works \cite{lippi2022enabling,lippi2022augment} and we refer to the quality measures \textit{\% any} and \textit{\% all}, where the former measures if any of the proposed plans results in a successful traversal from start to goal observation, while \textit{\% all} measure requires that all proposed paths are correct.  
Since our objective is to only output correct plans, we focus on the \textit{\% all} measure. 
We evaluate the performance by randomly selecting $1000$ different start and goal observations from a novel holdout set and, in case of harvesting, by checking that the path is feasible, e.g., the path is 
unfeasible if bunches are required to be moved from box to vine or from vine to vine to reach the goal observation. 
In the following, we access to the true system states for evaluation purposes only. 
 Unless specified otherwise, we use $c_{\max}=20$. 

\begin{figure}
    \centering
    \includegraphics[width=\linewidth]{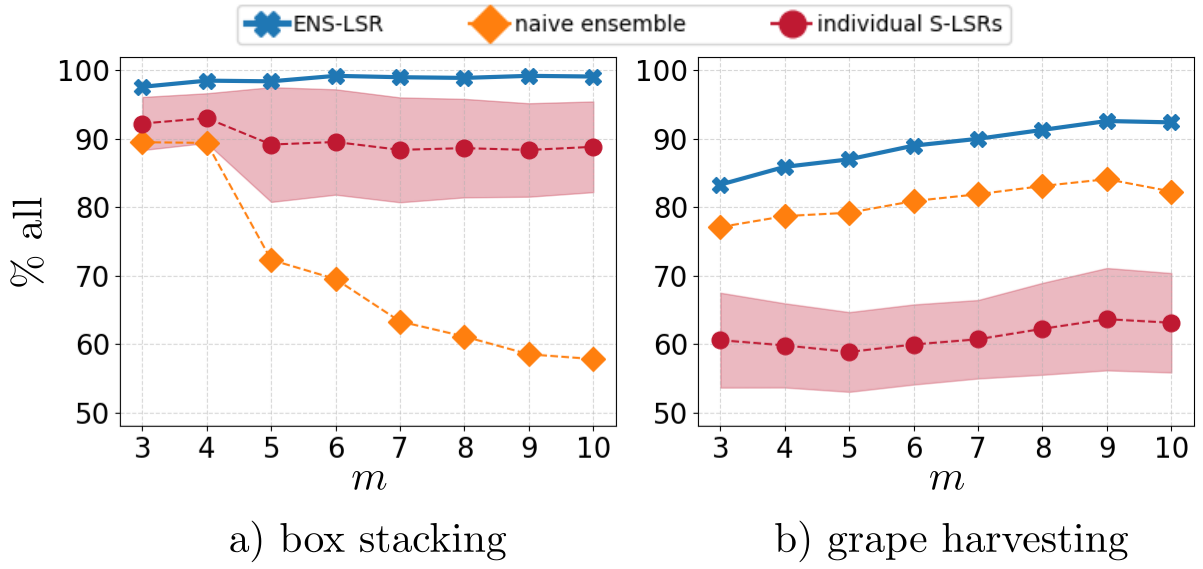}
\vspace{-20pt}
    \caption{Performance comparison of our ENS-LSR (in blue) with the individual models (in red, mean and variance are shown) and with the naive ensemble approach. Results for the box stacking task (left) and grape harvesting task (right) are reported. 
    }
    \label{fig:exe1_simu_results}
\end{figure}

\subsection{ENS-LSR via different mapping modules}
To answer question 1), we considered an ensemble of ten S-LSRs for each simulated task obtained by training ten different mapping modules. These were generated    
by randomly selecting $85\%$ of the training data. 
Results in terms of {\textit{\% all}} are reported in Fig. \ref{fig:exe1_simu_results}a for the box stacking task and in Fig. \ref{fig:exe1_simu_results}b for the grape harvesting task. In particular, the figure shows the performance of ENS-LSR (in blue) by increasing the number $m$ of S-LSRs from three to ten as well as it reports the performance achieved by the individual S-LSRs (in red) as well as by the naive ensemble approach (in orange) which outputs all possible VAPs, i.e., with $\mathcal{P}^*=\mathcal{P}$, as discussed in Sec.~\ref{sec:enslsr}. 

In both tasks, ENS-LSR significantly outperforms both the individual models and the naive ensemble approach. 
In particular, for the box stacking task, we can notice that simply combining all plans (naive approach) results in decreasing  performance as the number $m$ of S-LSRs increases.  In detail, it achieves $89.5\%$ with $m=3$ and $57.8\%$ with $m=10$. This is motivated by the fact that the higher $m$, 
the higher the likelihood that at least one wrong plan is suggested. 
As far as the individual models are concerned, we report both  mean (dots) and variance (shadow) of the performance obtained by utilizing $m$ S-LSRs. The figure highlights the significant variance arising from the individual models. 
 A slight performance drop can also be observed with the addition of the fifth S-LSR that only reaches $73.7\%$.
In contrast, 
our ensemble model ENS-LSR is not affected by the addition of this S-LSR and already scores $97.6\%$ with $m=3$ and  achieves $99.1\%$ with $m=10$.

Regarding the grape harvesting task,  the individual S-LSRs only  achieve about $60\%$ performance on average for all numbers of S-LSRs   
since they are often unable to find a path. This is overcome by the naive ensemble method, where  only one of the S-LSRs needs to find a path, reaching $77.1$ with $m=3$. 
 However,  also in this task, a performance decrease is recorded with the naive approach as the number of S-LSRs increases and  a model underperforming compared to the average is included (here, the decrease occurs with $m=10$). 
Our method not only outperforms the naive approach and the individual S-LSRs, achieving  $83.3\%$ with $m=3$ and $92.6\%$ with $m=9$, but is also robust when low performing models  are included, obtaining  $92.4\%$  with $m=10$.

\subsection{ENS-LSR via different LSR hyperparameters}
We investigated question 2) by considering an ensemble of ten S-LSRs differing with respect to the LSR hyperparameter $c_{\max}$, while employing the same mapping module. In detail, we studied  \mbox{$c_{\max}\in\{1, 10, 20, 30, 40, 50, 60, 70, 80, 90\}$}.

\begin{table*}[t]
\centering
\resizebox{\textwidth}{!}{\begin{tabular}{|l|c|c|c|c|c|c|c|c|c|c|c|}
\hline
{Stacking}  & {ENS-LSR} & $c_{\max}=1$ & $c_{\max}=10$ & $c_{\max}=20$ & $c_{\max}=30$ & $c_{\max}=40$ & $c_{\max}=50$ & $c_{\max}=60$ & $c_{\max}=70$ & $c_{\max}=80$ & $c_{\max}=90$ \\ \hline
{\textit{\% all}}   & \boldmath$98.8$ & $93.4$ & $96.4$ & $96.0$ & $94.5$ & $91.9$ & $88.5$ & $84.0$ & $80.3$ & $78.0$ & $75.4$ \\ \hline 
 {\textit{\% $\exists$ path}}   & \boldmath$100.0$ & \boldmath$100.0$ & $97.6$ & $96.9$ & $95.3$ & $92.9$ & $89.7$ & $85.2$ & $81.4$ & $79.1$ & $76.5$ \\ \hline 
\hline
\hline
{Harvesting}  & {ENS-LSR} & $c_{\max}=1$ & $c_{\max}=10$ & $c_{\max}=20$ & $c_{\max}=30$ & $c_{\max}=40$ & $c_{\max}=50$ & $c_{\max}=60$ & $c_{\max}=70$ & $c_{\max}=80$ & $c_{\max}=90$ \\ \hline 
{\textit{\% all}}   & \boldmath$69.6$ & $11.0$ & $57.6$ & $56.7$ & $54.8$ & $51.9$ & $48.2$ & $44.6$ & $40.9$ & $39.2$ & $38.5$ \\ \hline 
{\textit{\% $\exists$ path}}  & \boldmath$98.8$ & $98.1$ & $76.5$ & $65.1$ & $59.4$ & $54.4$ & $48.8$ & $44.7$ & $41.0$ & $39.3$ & $38.7$ \\ \hline 
\end{tabular}}
\vspace{-1pt}
\caption{Results in terms of \emph{\% all} performance and percentage of times a solution is found when considering 
different $c_{\max}$ hyperparameters. Both simulation tasks are reported (box stacking on top and grape harvesting on the bottom). Best results in bold.  } 
\label{tab:cmaxsimu_results}
\end{table*}
Table \ref{tab:cmaxsimu_results} summarizes the results of this analysis for both simulation tasks (stacking on top and harvesting on the bottom). 
Specifically, the performance index \textit{\% all} (second row of each block) and the percentage of times a VAP is found, denoted as \textit{\% $\exists$ path} (third row of each block), are reported for our ENS-LSR (second column) and for the individual S-LSRs (third to last column).
Results show that ENS-LSR outperforms any individual S-LSR in both tasks with regard to both metrics. More in detail, for the box stacking task, ENS-LSR achieves \textit{\% all} equal to $98.8\%$ outperforming the best individual model by $2.4\%$. Coherently  with results in \cite{lippi2022enabling}, best individual S-LSRs are obtained with $c_{\max}$  equal to $10$ and $20$ and a  performance decrease is obtained with higher $c_{\max}$ values, since more disconnected graphs are built, as well as with $c_{\max}=1$, since the method is not able to eliminate outliers in the training data. 
  The \textit{\% $\exists$ path} metric reaches maximum value with ENS-LSR, i.e., $100\%$, while for the individual S-LSRs, it starts from $100\%$ in case of a connected graph with $c_{\max}=1$ and then decreases as $c_{\max}$ increases. This is motivated by the fact that more disconnected graph components mean more risk of no path existing  from  start to goal nodes.

As far as the grape harvesting task is concerned, similar trends can be observed with ENS-LSR outperforming any 
individual S-LSR by at least $>10\%$ for \textit{\% all} and achieving $69.6\%$. Regarding the existence of a  solution, results show that ENS-LSR is able to plan a path most of the times, with \textit{\% $\exists$ path} equal to $98.8\%$, while it decreases for the individual S-LSRs as $c_{\max}$ increases,  dropping to $38.7\%$ with \mbox{$c_{\max}=90$}. 
Note that although $c_{\max}=1$ forces a single connected component, the corresponding S-LSR  only achieves $98.1\%$ for \textit{\% $\exists$ path} since it produces paths of zero length in $1.9\%$ of cases,  meaning that it incorrectly assumes that  starting and goal states have the same underlying state.

Overall, we can summarize that our ENS-LSR is able to improve performance over individual S-LSRs both with different 
mapping module, where the maximum enhancement is achieved, and with different LSR $c_{\max}$ hyperparameter, making its tuning even easier. Example visual action plans for both simulations are shown in the accompanying video.

\subsection{Similarity measure comparison}

 As for question 3), we analyzed  the influence of two additional action similarity measures and a node one and we conducted an ablation study on what part of Algorithm~\ref{alg::ensemble} is most beneficial to ENS-LSR.

Regarding the action similarity measures, first, we considered a simple baseline  based on the euclidean distance with action plans of same length, namely:
\begin{equation}
    s^u_{eucl}(P^u_{i,j},P^u_{k,l}) = -\|P^u_{i,j}-P^u_{k,l}\|, 
    \label{eq:dal2}
\end{equation}
which is $0$ when the plans are the same, and negative otherwise. 
In case the plans had different lengths, we assigned  no euclidean similarity measure. 
Next, we resorted to the edit similarity \cite{BILLE2005217}, which is commonly employed for comparing strings. In particular, this is based on counting the number of operations (insertion, deletion, or substitution) that are required to transform one plan into the other and weighting them according to given costs. To define if two actions $u_i$ and $u_j$ were the same, we required their euclidean distance to be lower than a threshold, i.e. $\|u_i-u_j\|<\tau$ where $\tau$ was set to $0.5$ in our tests. We employed the algorithm for the edit distance based on dynamic programming \cite{BILLE2005217} and considered costs for insertion and deletion  equal to $0.5$ and $1$, respectively. As far as the substitution cost is concerned, we set it equal 
As for the euclidean measure, we used as similarity measure the opposite of the distance.

As far as the node similarity is concerned, we considered a baseline where the Jaccard similarity between \emph{individual} compositions of nodes in the plans is computed and then aggregated for all nodes. 
In case of plans with different lengths, we assigned no similarity measure. We refer to this baseline as $s^n_{indiv}$ and obtain it as: 
\begin{equation}
    s^n_{indiv}(P_{i,j}^z,P_{k,l}^z)=  \sum_{t=1}^{n} \frac{|C_{i,j,t} \cap C_{k,l,t}|}{|C_{i,j,t}\cup C_{k,l,t}|} 
\end{equation}
with 
$C_{i,j,t}$ and $C_{k,l,t}$ the compositions of the $t$\ts th node in the plans  $P_{i,j}^z$ and $P_{k,l}^z$, respectively.

\begin{figure}
    \centering
    \includegraphics[width=\linewidth]{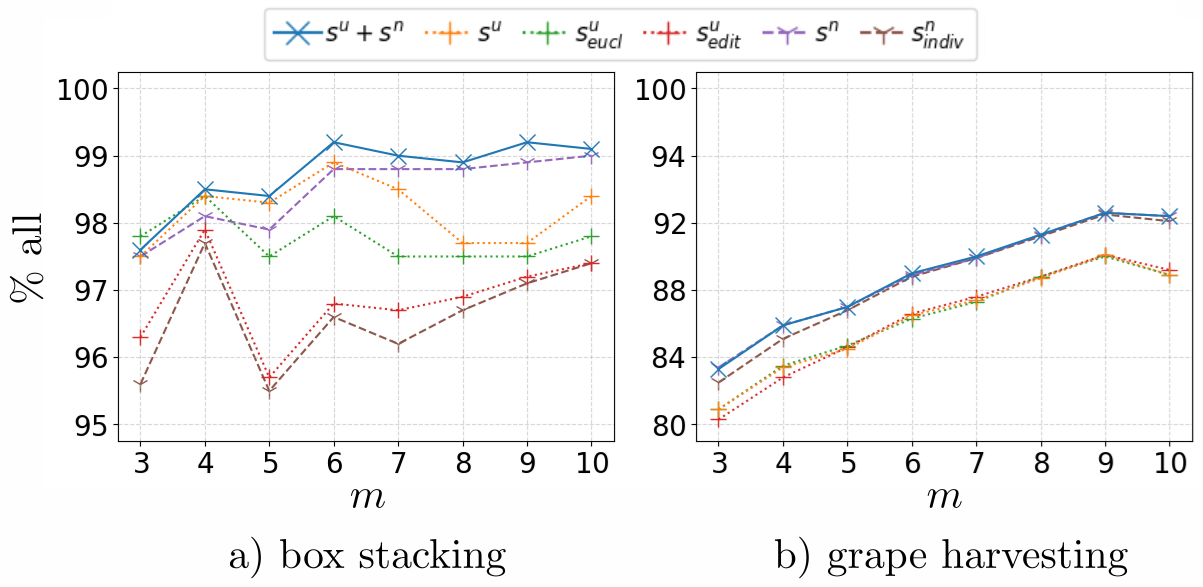}
\vspace{-20pt}
    \caption{Comparison with different similarity measures on the box stacking (left) and grape harvesting (right) tasks.
    }
    \label{fig:ablation_distance}
\end{figure}

We compared the performance of the ensemble when using each similarity measure individually, i.e., by adding to the set $\mathcal{S}$ in line \ref{lst:line:addtos} of Algorithm~\ref{alg::ensemble} the respective measure.  
Results are shown in  Fig. \ref{fig:ablation_distance} for both simulation tasks (box stacking task on the left and grape harvesting task on the right). In particular, our ENS-LSR results, using the overall similarity, are shown in blue, the results with individual measures of action similarity based on cosine, $s^u$, euclidean distance, $s^u_{eucl}$, and edit distance, $s^u_{edit}$, are depicted in orange, green, and red, respectively, while the results with  measures of node similarity based on the path Jaccard similarity, $s^n$, and on the individual node-based Jaccard similarity, $s^n_{indiv}$, are represented in purple and brown, respectively.

For the box stacking task, we observe that all similarity measures lead to very high performance \textit{\% all}, but in general the ones requiring equal path lengths (i.e., $s^u_{eucl}$ and $s^n_{indiv}$) perform worse, while $s^n$ leads to more stable results compared to the action-based measures. Moreover, the figure shows that $s^u$ and $s^n$ generally lead to the best performance as action- and the node-based similarity measures, respectively. However, the overall similarity measure (i.e., $s^n+s^u$) outperforms all others and remains robust even if any individual similarity measure underperforms.
For the grape harvesting task, 
 a small distinction is recorded between methods that require equal path length and those that do not. This is mainly due to the fact that the S-LSRs generate noticeably fewer plans compared to the stacking task and may not find any plans in some cases, thus reducing the frequency of situations where plans of differing lengths need to be compared. 
However, we can notice  a clear performance difference between node-based similarity measures and action-based ones. Specifically, the former outperforms the latter by approximately $3\%$. 
This can be motivated by the fact that node-based similarity measures have much more fine-grained information 
compared to the action-based ones 
which is beneficial for a more accurate comparison. 
Finally, the figure shows that  $s^n+s^u$  does not lead to any undesirable performance drop for $m=10$ compared to  
$s^n_{indiv}$.

\section{Folding results}\label{sec:exp_fold}
\begin{figure}
    \centering
    \includegraphics[width=\linewidth]{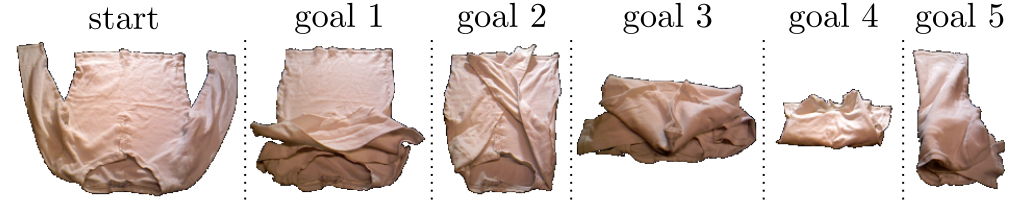}
\vspace{-15pt}
    \caption{Start and goal configurations for the real-world folding task.}
    \label{fig:folds}
\end{figure}

We validated the effectiveness of ENS-LSR in a real-world T-shirt folding task as in \cite{lippi2022augment}. This task consists of a start state, with the T-shirt spread on the table, and five goal states,  as shown in Fig. \ref{fig:folds}. Actions are expressed as pick-and-place operations as in the simulation tasks and are not reversible, leading to directed LSRs.  All execution videos and the dataset are available on the paper website\usefootref{fn:website}. 
To build the ENS-LSR we  leveraged the insights from the simulation tasks, where  a variety of mapping modules yielded to greater performance improvements compared to a variety of LSR hyperparameters only. We, therefore, constructed an ENS-LSR consisting of ten distinct mapping modules  and three different $c_{\max}$ values - specifically, $1$, $5$, and $10$, leading to $m=30$. 
We set the minimum number of nodes in a cluster to $4$ for all S-LSRs. At planning time, based on Algorithm~\ref{alg::ensemble}, the ENS-LSR  takes both the current T-shirt observation and a specified goal configuration as input, and produces one or more visual action plans as output $\mathcal{P}^*$. The first action of the plan is then executed with a Baxter robot, as shown in the accompanying video and on the project website\usefootref{fn:website}. After each action, a replanning step is made using the current state as start configuration. This procedure is repeated until ENS-LSR determines that the current state has the same underlying state as the desired goal state or when no path is found. It is worth noticing that, since ENS-LSR aims to suggest plans which are all correct, it eliminates the need for human selection as in \cite{lippi2022enabling, lippi2022augment}, but  simply  randomly selects and executes a VAP in $\mathcal{P}^*$.

\begin{table}[]
\resizebox{\linewidth}{!}{
\begin{tabular}{|l|l|l|l|l|l|}
\hline
Framework & Fold 1 & Fold 2 & Fold 3 & Fold 4 & Fold 5 \\ \hline
ENS-LSR  &   \boldmath$5/5$     &     \boldmath$5/5$   &   \boldmath$4/5$     &  \boldmath$5/5$   &  $3/5$      \\ \hline
ACE-LSR   &    $4/5$    &   \boldmath$5/5$     &    $3/5$    &   $4/5$     &     \boldmath$4/5$   \\ \hline
\end{tabular}
}
\vspace{-1pt}
\caption{Performance results on the T-shirt folding task with ENS-LSR and ACE-LSR  on five different folds, each repeated five times. Best results in bold.  }
\label{tab:exe_fold}
\end{table}

We repeated the  execution  to each goal configuration five times as in our previous works and reported the system performance in Table \ref{tab:exe_fold}, where we compared it with our latest framework \cite{lippi2022augment}, i.e., ACE-LSR, which implements a paradigm for dealing with data scarcity. We employed for both models a dataset composed of $562$ tuples, which is $50\%$ of the training data used in \cite{lippi2022enabling}. 
Results show that ENS-LSR is generally more robust than ACE-LSR, successfully executing folds $1$, $2$, and $4$ at all times, and achieving overall performance of $88\%$, which  outperforms by $8\%$ the previously top-performing ACE-LSR model. 
A slight performance decrease is only recorded with 
Fold $5$, where ENS-LSR occasionally misses the final step that involves picking up a small T-shirt edge, as shown on the website\usefootref{fn:website}.

\section{Conclusions}

In this work we presented a novel ensemble algorithm for visual action planning. The method relies on multiple latent space roadmaps-based systems, 
which can be obtained either by different mapping modules or LSR  hyperparameters. Given start and goal observations, a selection of the plans to output is  made  on the basis of action- and node-based similarity measures. We analyzed the performance on two simulation tasks: 
the box stacking task from \cite{lippi2022enabling} and a new grape harvesting task. In both cases, the ensemble model significantly   outperformed the individual models, reaching $99.2\%$ and $92.6\%$ performance in terms of path correctness compared to $96.4\%$ and $74.9\%$ of the best performing individual models, respectively. 
For final validation, we compared the performance of the ensemble algorithm with \cite{lippi2022augment} on a real-world folding task, where the ensemble improved the performance by $8\%$ for a total of  $88\%$ success rate. In future work, we aim to leverage different latent spaces also to facilitate the transfer of knowledge across a range of tasks.


\AtNextBibliography{\scriptsize}

\printbibliography

\end{document}